\documentclass{article}

\usepackage[utf8]{inputenc}
\usepackage[table]{xcolor}
\usepackage{nips15submit_e,times}
\usepackage[square,sort,numbers]{natbib}
\usepackage[small,tight]{bibhacks}
\usepackage{hyperref}
\usepackage{url}
\usepackage{pgfplots}
\usepackage{amsmath}
\usepackage{amssymb}
\usepackage{bm}
\usepackage{ifthen}
\usepackage{pbox}
\usepackage{floatrow}
\usepackage{sansmath}

\usetikzlibrary{arrows}
\usetikzlibrary{positioning}
\usepgfplotslibrary{external}

\title{Generative Image Modeling Using Spatial LSTMs}

\author{
Lucas Theis \\
University of Tübingen \\
72076 Tübingen, Germany \\
\texttt{lucas@bethgelab.org} \\
\And
Matthias Bethge \\
University of Tübingen \\
72076 Tübingen, Germany \\
\texttt{matthias@bethgelab.org} \\
}

\newfloatcommand{capbtabbox}{table}[][\FBwidth]

\nipsfinalcopy 

\begin{document}
	\maketitle

	\begin{abstract}
	Modeling the distribution of natural images is challenging, partly because of
	strong statistical dependencies which can extend over hundreds of pixels.
	Recurrent neural networks have been successful in capturing long-range dependencies
	in a number of problems but only recently have found their way into generative image
	models. We here introduce a recurrent image model based on multi-dimensional long short-term
	memory units which are particularly suited for image modeling due to their
	spatial structure. Our model scales to images of arbitrary size and its likelihood is
	computationally tractable. We find that it outperforms the state of the art in quantitative
	comparisons on several image datasets and produces promising results when used for texture
	synthesis and inpainting.
	\end{abstract}

	\section{Introduction}
		The last few years have seen tremendous progress in learning useful image representations
		\cite{Donahue:2014}. While early successes were often achieved through the use of generative
		models \cite[e.g.,][]{Hinton:2006,Lee:2009,Ranzato:2011}, recent breakthroughs were mainly
		driven by improvements in supervised techniques \cite[e.g.,][]{Krizhevsky:2012,Simonyan:2015}.
		Yet unsupervised learning has the potential to tap into the much larger source of unlabeled
		data, which may be important for training bigger systems capable of a more
		general scene understanding. For example, multimodal data is abundant but often unlabeled,
		yet can still greatly benefit unsupervised approaches \cite{Srivastava:2014}.

		Generative models provide a principled approach to unsupervised learning.
		A perfect model of natural images would be able to optimally predict parts of an image given other parts
		of an image and thereby clearly demonstrate a form of scene understanding.
		When extended by labels, the Bayesian framework can be used to perform semi-supervised learning
		in the generative model \cite{Ngiam:2011,Kingma:2014b} while it is less clear how
		to combine other unsupervised approaches with discriminative learning.
		Generative image models are also useful in more traditional applications such as
		image reconstruction \cite{Roth:2009,Zoran:2011,Sohl-Dickstein:2015} or compression \cite{VanDenOord:2014b}.

		Recently there has been a renewed strong interest in the development of generative image models
		\cite[e.g.,][]{VanDenOord:2014b,Kingma:2014a,Uria:2014,Gregor:2014,Goodfellow:2014,Ranzato:2014,Gregor:2015,Sohl-Dickstein:2015,Li:2015,Denton:2015}.
		Most of this work has tried to bring to bear the flexibility of deep neural networks on the
		problem of modeling the distribution of natural images. One challenge in this endeavor is to
		find the right balance between tractability and flexibility. The present article contributes
		to this line of research by introducing a fully tractable yet highly flexible image model.

		Our model combines multi-dimensional recurrent neural networks \cite{Graves:2009} with
		mixtures of experts. More specifically, the backbone of our model is formed by a spatial variant of
		\textit{long short-term memory} (LSTM) \cite{Hochreiter:1997}.
		One-dimensional LSTMs have been particularly successful in modeling text and speech
		\cite[e.g.,][]{Sundermeyer:2010,Sutskever:2014}, but have also been used to model the progression of frames in video \cite{Srivastava:2014} and very
		recently to model single images \cite{Gregor:2015}. In contrast to earlier work on modeling
		images, here we use \textit{multi-dimensional LSTMs} \cite{Graves:2009} which naturally lend themselves to the task
		of generative image modeling due to their spatial structure and ability to capture long-range
		correlations.

		To model the distribution of pixels conditioned on the hidden states of the neural network, we use
		\textit{mixtures of conditional Gaussian scale mixtures} (MCGSMs) \cite{Theis:2012a}.
		This class of models can be viewed as a generalization of Gaussian mixture models, but
		their parametrization makes them much more suitable for natural images. By treating
		images as instances of a stationary stochastic process, this model allows us to sample and capture the
		correlations of arbitrarily large images.

	\section{A recurrent model of natural images}
		\label{sec:ride}
		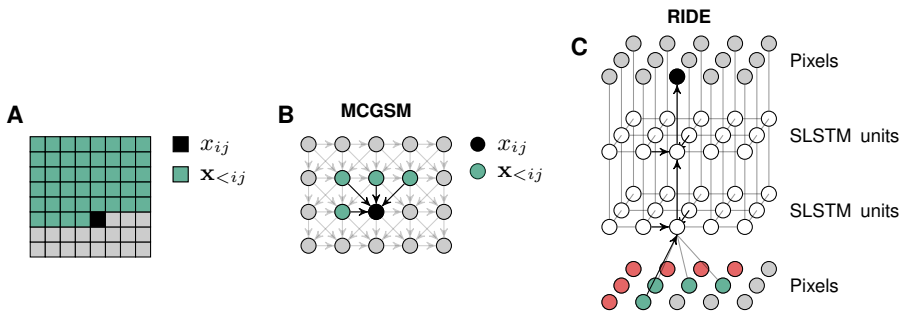
\begin{figure}
			\centering
			\tikzset{hunit/.style={circle,draw,minimum size=0.2cm,inner sep=0cm,fill=white}}
\tikzset{vunit/.style={circle,draw,minimum size=0.2cm,inner sep=0cm,fill=black!20!white}}
\tikzset{aunit/.style={circle,minimum size=0.2cm,inner sep=0cm,fill=blue!40!green,opacity=.5}}
\tikzset{munit/.style={circle,minimum size=0.2cm,inner sep=0cm,fill=red,opacity=.5}}
\tikzset{tunit/.style={circle,minimum size=0.2cm,inner sep=0cm,fill=black}}
\tikzset{connection/.style={-,black,opacity=.4}}
\tikzset{arrow/.style={-stealth',opacity=1}}
\tikzset{pixel/.style={rectangle,draw,minimum size=0.2cm,inner sep=0cm,fill=black!20!white}}
\tikzset{parent/.style={rectangle,draw,minimum size=0.2cm,inner sep=0cm,fill=blue!40!green,opacity=.5}}

\begin{tikzpicture}
  \def\zoffset{.16};
  \def\xoffset{.45};
  \def\yoffset{.22};


  \node[align=left, text width=1.5cm] at (7 * \xoffset, 3 + \yoffset) {\scriptsize\textsf{Pixels}};
  \node[align=left, text width=1.5cm] at (7 * \xoffset, 2 + \yoffset) {\scriptsize\textsf{SLSTM units}};
  \node[align=left, text width=1.5cm] at (7 * \xoffset, 1 + \yoffset) {\scriptsize\textsf{SLSTM units}};
  \node[align=left, text width=1.5cm] at (7 * \xoffset, 0 + \yoffset) {\scriptsize\textsf{Pixels}};

  \node at (2 * \xoffset + \zoffset, 3.8) {\scriptsize\textsf{\textbf{RIDE}}};

  \foreach \z in {2, 1, 0} {
    \foreach \x in {0, 1, 2, 3, 4} {
      \node[draw,vunit] (v0\x\z) at (\x * \xoffset + \z * \zoffset, \z * \yoffset) {};
      \node[draw,hunit] (h0\x\z) at (\x * \xoffset + \z * \zoffset, \z * \yoffset + 1) {};
      \node[draw,hunit] (h1\x\z) at (\x * \xoffset + \z * \zoffset, \z * \yoffset + 2) {};
      \node[draw,vunit] (v1\x\z) at (\x * \xoffset + \z * \zoffset, \z * \yoffset + 3) {};

      \draw [connection] (h0\x\z.north) to (h1\x\z.south);
      \draw [connection] (h1\x\z.north) to (v1\x\z.south);

      \ifthenelse{\z > 0}{
        \draw [connection] (\x * \xoffset + \z * \zoffset, \z * \yoffset + 1) to (\x * \xoffset + \z * \zoffset - \zoffset, \z * \yoffset - \yoffset + 1);
        \draw [connection] (\x * \xoffset + \z * \zoffset, \z * \yoffset + 2) to (\x * \xoffset + \z * \zoffset - \zoffset, \z * \yoffset - \yoffset + 2);}{};
    }

    \ifthenelse{\z = 1}{
     \draw [connection,arrow] (2 * \xoffset + \z * \zoffset, \z * \yoffset + 1) to (2 * \xoffset + \z * \zoffset - \zoffset, \z * \yoffset - \yoffset + 1);
     \draw [connection,arrow] (2 * \xoffset + \z * \zoffset, \z * \yoffset + 2) to (2 * \xoffset + \z * \zoffset - \zoffset, \z * \yoffset - \yoffset + 2);
      \foreach \x in {1, 2, 3} {
        \node [draw,aunit] at (\x * \xoffset + 1 * \zoffset, 1 * \yoffset) {};
      }
    }{}

    \ifthenelse{\z = 0}{
      \node[draw,aunit] at (1 * \xoffset, 0) {};

      \foreach \x in {1, 2, 3}
        \draw [connection] (v0\x1) to (h020.south);
      \draw [connection,arrow] (v010) to (h020.south);

      \node [draw,tunit] at (2 * \xoffset + 0 * \zoffset, 0 * \yoffset + 3) {};
    }{}

    \ifthenelse{\z = 3 \OR \z = 2}{
      \foreach \x in {0, 1, 2, 3} {
        \node [draw,munit] at (\x * \xoffset + \z * \zoffset, \z * \yoffset) {};
      }
    }{}

    \ifthenelse{\z = 1 \OR \z = 0}{
      \foreach \x in {0} {
        \node [draw,munit] at (\x * \xoffset + \z * \zoffset, \z * \yoffset) {};
      }
    }{}

    \foreach \x in {0, 1, 2, 3} {
      \pgfmathparse{\x + 1};
      \pgfmathtruncatemacro\r{\pgfmathresult};
      \draw [connection] (h0\x\z.east) to (h0\r\z.west);
      \draw [connection] (h1\x\z.east) to (h1\r\z.west);
    }
  }
  \draw [connection,arrow] (h120.north) to (v120.south);
  \draw [connection,arrow] (h020.north) to (h120.south);
  \draw [connection,arrow] (h010.east) to (h020.west);
  \draw [connection,arrow] (h110.east) to (h120.west);


  \begin{scope}[xshift=-4cm,yshift=1.2cm]
    \def\zoffset{0};
    \def\xoffset{.45};
    \def\yoffset{.45};

    \node[tunit] (l3) at (5 * \xoffset, 2 * \yoffset) {};
    \node[vunit] (l4) at (5 * \xoffset, 1.2 * \yoffset) {};
    \node[aunit] at (5 * \xoffset, 1.2 * \yoffset) {};
    \node[right=0.03cm of l3,yshift=-.04cm] {\footnotesize $x_{ij}$};
    \node[right=0.03cm of l4,yshift=-.04cm] {\footnotesize $\mathbf{x}_{<ij}$};

    \node at (2 * \xoffset, 3 * \yoffset) {\scriptsize\textsf{\textbf{MCGSM}}};

    \foreach \z in {2, 1, 0, -1} {
      \foreach \x in {0, 1, 2, 3, 4}
        \node[draw,vunit] (v\x\z) at (\x * \xoffset + \z * \zoffset, \z * \yoffset) {};

      \ifthenelse{\z = 1}{
        \foreach \x in {1, 2, 3}
          \node [draw,aunit] at (\x * \xoffset + 1 * \zoffset, 1 * \yoffset) {};
      }{}
    }
    \node [draw,aunit] at (1 * \xoffset, 0) {};
    \node [draw,tunit] at (2 * \xoffset, 0) {};

    \foreach \z in {1, 0, -1}
      \foreach \x in {0, 1, 2, 3, 4} {
        \pgfmathparse{\z + 1};
        \pgfmathtruncatemacro\r{\pgfmathresult};
        \draw [connection,arrow,opacity=.2] (v\x\r) to (v\x\z) {};
      }

    \foreach \z in {2, 1, 0, -1}
      \foreach \x in {0, 1, 2, 3} {
        \pgfmathparse{\x + 1};
        \pgfmathtruncatemacro\r{\pgfmathresult};
        \draw [connection,arrow,opacity=.2] (v\x\z) to (v\r\z) {};
      }

    \foreach \z in {1, 0, -1}
      \foreach \x in {0, 1, 2, 3} {
        \pgfmathparse{\x + 1};
        \pgfmathtruncatemacro\r{\pgfmathresult};
        \pgfmathparse{\z + 1};
        \pgfmathtruncatemacro\s{\pgfmathresult};
        \draw [connection,arrow,opacity=.2] (v\x\s) to (v\r\z) {};
        \draw [connection,arrow,opacity=.2] (v\r\s) to (v\x\z) {};
      }

    \draw [connection,arrow] (v10) to (v20);
    \draw [connection,arrow] (v11) to (v20);
    \draw [connection,arrow] (v21) to (v20);
    \draw [connection,arrow] (v31) to (v20);
  \end{scope}

  \node at (-7.9cm,2.5cm) {\small\textsf{\textbf{A}}};
  \node at (-4.3cm,2.5cm) {\small\textsf{\textbf{B}}};
  \node at (-0.4cm,3.4cm) {\small\textsf{\textbf{C}}};

  \begin{scope}[xshift=-7.6cm,yshift=.7cm]
    \node[pixel,fill=black] (l1) at (9.5 * 0.2cm, 7 * 0.2cm) {};
    \node[pixel] (l2) at (9.5 * 0.2cm, 5 * 0.2cm) {};
    \node[parent] at (9.5 * 0.2cm, 5 * 0.2cm) {};
    \node[right=0.03cm of l1,yshift=-.04cm] {\footnotesize $x_{ij}$};
    \node[right=0.03cm of l2,yshift=-.04cm] {\footnotesize $\mathbf{x}_{<ij}$};

    \foreach \y in {0, 1, 2, 3, 4, 5, 6, 7}
      \foreach \x in {0, 1, 2, 3, 4, 5, 6, 7}
        \node[pixel] at (\x * 0.2cm, \y * 0.2cm) {};
    \foreach \y in {3, 4, 5, 6, 7}
      \foreach \x in {0, 1, 2, 3, 4, 5, 6, 7}
        \node[parent] at (\x * 0.2cm, \y * 0.2cm) {};
    \foreach \x in {0, 1, 2, 3}
      \node[parent] at (\x * 0.2cm, 2 * 0.2cm) {};
    \node[pixel,fill=black] at (4 * 0.2cm, 2 * 0.2cm) {};
  \end{scope}
\end{tikzpicture}
			\caption{(\textbf{A}) We factorize the distribution of images such that the prediction
				of a pixel (black) may depend on any pixel in the upper-left green region. (\textbf{B}) A graphical model representation of an MCGSM with a
				causal neighborhood limited to a small region. (\textbf{C}) A visualization of our recurrent image model with two layers of spatial LSTMs.
				The pixels of the image are represented twice and some
				arrows are omitted for clarity. Through feedforward connections, the prediction of a pixel depends
				directly on its neighborhood (green), but through recurrent connections it has
				access to the information in a much larger region (red).}
			\label{fig:rim}
		\end{figure}
		In the following, we first review and extend the MCGSM \cite{Theis:2012a} and
		multi-dimensional LSTMs \cite{Graves:2009} before explaining how to combine them into
		a recurrent image model. Section~\ref{sec:experiments} will demonstrate
		the validity of our approach by evaluating and comparing the model on a number of image datasets.

		\subsection{Factorized mixtures of conditional Gaussian scale mixtures}
		\label{sec:mcgsm}
		One successful approach to building flexible yet tractable generative models has been to
		use fully-visible belief networks \cite{Neal:1992,Larochelle:2011}. To apply
		such a model to images, we have to give the pixels an ordering and specify the
		distribution of each pixel conditioned on its parent pixels. Several parametrizations
		have been suggested for the conditional distributions in the context of natural images \cite{Domke:2008,Hosseini:2010,Theis:2012a,Uria:2013,Uria:2014}.
		We here review and extend the work of Theis et al. \cite{Theis:2012a} who proposed
		to use \textit{mixtures of conditional Gaussian scale mixtures} (MCGSMs).

		Let $\mathbf{x}$ be a grayscale image patch and $x_{ij}$ be the intensity of the pixel at location
		$ij$. Further, let $\mathbf{x}_{<ij}$ designate the set of pixels $x_{mn}$ such that $m < i$
		or $m = i$ and $n < j$ (Figure~\ref{fig:rim}A). Then
		\begin{align}
			\textstyle
			p(\mathbf{x}; \bm{\theta}) = \prod_{i,j} p(x_{ij} \mid \mathbf{x}_{<ij}; \bm{\theta})
			\label{eq:factorization}
		\end{align}
		for the distribution of any parametric model with parameters $\bm{\theta}$. Note that this
		factorization does not make any independence assumptions but is simply an
		application of the probability chain rule. Further note that the conditional distributions
		all share the same set of parameters. One way to improve the
		representational power of a model is thus to endow each conditional distribution with its own set of parameters,
		\begin{align}
			\textstyle
			p(\mathbf{x}; \left\{ \bm{\theta}_{ij} \right\}) = \prod_{i,j} p(x_{ij} \mid \mathbf{x}_{<ij}; \bm{\theta}_{ij}).
			\label{eq:factorization2}
		\end{align}
		Applying this trick to mixtures of Gaussian scale mixtures (MoGSMs) yields the MCGSM \cite{Theis:2011}.
		Untying shared parameters can drastically increase the number of parameters. For images, it
		can easily be reduced again by adding assumptions. For example, we can limit
		$\mathbf{x}_{<ij}$ to a smaller neighborhood surrounding the pixel by making a Markov
		assumption. We will refer to the resulting set of parents as the pixel's \textit{causal neighborhood} (Figure~\ref{fig:rim}B).
		Another reasonable assumption is stationarity or shift invariance, in which case we only have
		to learn one set of parameters $\bm{\theta}_{ij}$ which can then be used at every pixel
		location. Similar to convolutions in neural networks, this allows the model to easily scale to images of arbitrary size.
		While this assumption reintroduces parameter sharing constraints into the model, the constraints are
		different from the ones induced by the joint mixture model.

		The conditional distribution in an MCGSM takes the form of a mixture of experts,
		\begin{align}
			p(x_{ij} \mid \mathbf{x}_{<ij}, \bm{\theta}_{ij})
			&= \sum_{c,s} \underbrace{p(c, s \mid \mathbf{x}_{<ij}, \bm{\theta}_{ij})}_\text{gate} \underbrace{p(x_{ij} \mid \mathbf{x}_{<ij}, c, s, \bm{\theta}_{ij})}_\text{expert},
		\end{align}
		where the sum is over mixture component indices $c$ corresponding to different covariances
		and scales $s$ corresponding to different variances.
		The gates and experts in an MCGSM are given by
		\begin{align}
			p(c, s \mid \mathbf{x}_{<ij})
			&\propto \textstyle \exp\left( \eta_{cs} - \frac{1}{2} e^{\alpha_{cs}} \mathbf{x}_{<ij}^\top \mathbf{K}_c \mathbf{x}_{<ij} \right),
			\label{eq:gates} \\
			p(x_{ij} \mid \mathbf{x}_{<ij}, c, s)
			&= \mathcal{N}(x_{ij}; \mathbf{a}_c^\top \mathbf{x}_{<ij}, e^{-\alpha_{cs}}),
			\label{eq:experts}
		\end{align}
		where $\mathbf{K}_c$ is positive definite. The number of parameters of an MCGSM still grows quadratically with
		the dimensionality of the causal neighborhood. To further reduce the number of parameters,
		we introduce a factorized form of the MCGSM with additional parameter sharing by replacing
		$\mathbf{K}_c$ with $\sum_n \beta_{cn}^2 \mathbf{b}_n\mathbf{b}_n^\top$.
		This \textit{factorized MCGSM} allows us to use larger neighborhoods and more mixture components.
		A detailed derivation of a more general version which also allows for multivariate pixels is given in
		Supplementary Section~1.

		\subsection{Spatial long short-term memory}
		In the following we briefly describe the \textit{spatial LSTM} (SLSTM), a special case of the
		multi-dimensional LSTM first described by Graves \& Schmidhuber \cite{Graves:2009}.
		At the core of the model are memory units $\mathbf{c}_{ij}$ and hidden units
		$\mathbf{h}_{ij}$. For each location $ij$ on a two-dimensional grid, the operations
		performed by the spatial LSTM are given by \\
		\begin{minipage}{0.5\textwidth}
			\begin{align*}
				\mathbf{c}_{ij} &=
					\mathbf{g}_{ij} \odot \mathbf{i}_{ij}
					+ \mathbf{c}_{i,j - 1} \odot \mathbf{f}_{ij}^c
					+ \mathbf{c}_{i - 1,j} \odot \mathbf{f}_{ij}^r, \\
				\mathbf{h}_{ij} &= \tanh\left( \mathbf{c}_{ij} \odot \mathbf{o}_{ij} \right),
			\end{align*}
		\end{minipage}
		\begin{minipage}{0.5\textwidth}
			\begin{align}
				\begin{pmatrix}
					\mathbf{g}_{ij} \\
					\mathbf{o}_{ij} \\
					\mathbf{i}_{ij} \\
					\mathbf{f}_{ij}^r \\
					\mathbf{f}_{ij}^c
				\end{pmatrix}
				=
				\begin{pmatrix}
					\tanh \\
					\sigma \\
					\sigma \\
					\sigma \\
					\sigma
				\end{pmatrix}
				T_{\mathbf{A},\mathbf{b}}
				\begin{pmatrix}
					\mathbf{x}_{<ij} \\
					\mathbf{h}_{i,j - 1} \\
					\mathbf{h}_{i - 1,j}
				\end{pmatrix},
				\label{eq:slstm}
			\end{align}
		\end{minipage} \\[.4cm]
		where $\sigma$ is the logistic sigmoid function, $\odot$ indicates a pointwise product,
		and $T_{\mathbf{A},\mathbf{b}}$ is an affine transformation which depends on the only
		parameters of the network $\mathbf{A}$ and $\mathbf{b}$. The gating units $\mathbf{i}_{ij}$ and $\mathbf{o}_{ij}$
		determine which memory units are affected by the inputs through $\mathbf{g}_{ij}$, and which memory states
		are written to the hidden units $\mathbf{h}_{ij}$. In contrast to
		a regular LSTM defined over time, each memory unit of a spatial LSTM has two preceding states
		$\mathbf{c}_{i,j-1}$ and $\mathbf{c}_{i-1,j}$ and two corresponding forget gates
		$\mathbf{f}_{ij}^c$ and $\mathbf{f}_{ij}^r$.

		\subsection{Recurrent image density estimator}
		We use a grid of SLSTM units to sequentially read relatively small neighborhoods of pixels
		from the image, producing a hidden vector at every pixel. The hidden states are then fed
		into a factorized MCGSM to predict the state of the corresponding pixel, that is,
		$p(x_{ij} \mid \mathbf{x}_{<ij}) = p(x_{ij} \mid \mathbf{h}_{ij})$.
		Importantly, the state of the hidden vector only depends on pixels in $\mathbf{x}_{<ij}$ and
		does not violate the factorization given in Equation~\ref{eq:factorization}. Nevertheless,
		the recurrent network allows this \textit{recurrent image density estimator} (RIDE) to use pixels of a much larger
		region for prediction, and to nonlinearly transform the pixels before applying the MCGSM. We
		can further increase the representational power of the model by stacking spatial LSTMs
		to obtain a deep yet still completely tractable recurrent image model (Figure~\ref{fig:rim}C).

		\subsection{Related work}
			Larochelle \& Murray \cite{Larochelle:2011} derived a tractable density estimator (NADE)
			in a manner similar to how the MCGSM was derived \cite{Theis:2012a}, but using restricted Boltzmann
			machines (RBM) instead of mixture models as a starting point. In contrast to the MCGSM,
			NADE tries to keep the weight sharing constraints induced by the RBM (Equation~\ref{eq:factorization}).
			Uria et al. extended NADE to real values \cite{Uria:2013} and introduced hidden layers to the
			model \cite{Uria:2014}. Gregor et al. \cite{Gregor:2014} describe a related
			autoregressive network for binary data which additionally allows for stochastic hidden units.

			Gregor et al. \cite{Gregor:2015} used one-dimensional LSTMs to generate images in a sequential
			manner (DRAW). Because the model was defined over Bernoulli variables,
			normalized RGB values had to be treated as probabilities, making a direct comparison with
			other image models difficult. In contrast to our model, the presence of stochastic latent variables
			in DRAW means that its likelihood cannot be evaluated but has to be approximated.

			Ranzato et al. \cite{Ranzato:2014} and Srivastava et al. \cite{Srivastava:2015} use
			one-dimensional recurrent neural networks to model videos, but recurrency is not used to
			describe the distribution over individual frames. Srivastava et al. \cite{Srivastava:2015}
			optimize a squared error corresponding to a Gaussian assumption, while Ranzato et
			al. \cite{Ranzato:2014} try to side-step having to model pixel intensities by quantizing image patches. In
			contrast, here we also try to solve the problem of modeling pixel intensities by using
			an MCGSM, which is equipped to model heavy-tailed as well as multi-modal distributions.

	\section{Experiments}
		\label{sec:experiments}
		RIDE was trained using stochastic gradient descent with a batch size of 50, momentum of 0.9, and a decreasing learning rate varying
		between 1 and $10^{-4}$.
		After each pass through the training set, the MCGSM of RIDE was finetuned using L-BFGS for
		up to 500 iterations before decreasing the learning rate. No regularization was used except for early stopping based on
		a validation set. Except where indicated otherwise, the recurrent model used a
		5 pixel wide neighborhood and an MCGSM with 32 components and 32
		quadratic features ($\mathbf{b}_n$ in Section~\ref{sec:mcgsm}). Spatial LSTMs were implemented using
		the Caffe framework \cite{jia:2014}. Where appropriate, we augmented the data by horizontal
		or vertical flipping of images.

		We found that conditionally whitening the data greatly sped up the training process of both
		models. Letting $\mathbf{y}$ represent a pixel and
		$\mathbf{x}$ its causal neighborhood, conditional whitening replaces these with
		\begin{align}
			\mathbf{\hat x} &= \textstyle\mathbf{C}_\mathbf{xx}^{-\frac{1}{2}} \left( \mathbf{x} - \mathbf{m}_\mathbf{x} \right), &
			\mathbf{\hat y} &= \textstyle \mathbf{W} (\mathbf{y} - \mathbf{C}_{\mathbf{yx}} \mathbf{C}_\mathbf{xx}^{-\frac{1}{2}} \mathbf{\hat x} - \mathbf{m}_\mathbf{y}), &
			\mathbf{W} &= (\mathbf{C}_\mathbf{yy} - \mathbf{C}_{\mathbf{yx}} \mathbf{C}_\mathbf{xx}^{-1} \mathbf{C}_{\mathbf{yx}}^\top)^{-\frac{1}{2}},
		\end{align}
		where
		$\mathbf{C}_{\mathbf{yx}}$ is the covariance of $\mathbf{y}$ and $\mathbf{x}$, and
		$\mathbf{m}_\mathbf{x}$ is the mean of $\mathbf{x}$.
		In addition to speeding up training, this variance normalization step helps to make the learning
		rates less dependent on the training data. When evaluating the conditional log-likelihood,
		we compensate for the change in variance by adding the log-Jacobian $\log |\det \mathbf{W}|$. Note that this preconditioning introduces a
		shortcut connection from the pixel neighborhood to the predicted pixel which is not shown in Figure~\ref{fig:rim}C.

		\subsection{Ensembles}
		Uria et al. \cite{Uria:2014} found that forming ensembles of their autoregressive model over
		different pixel orderings significantly improved performance. We here consider a simple trick to
		produce an ensemble without the need for training different models or to change training
		procedures. If $\mathbf{T}_k$ are linear transformations leaving the targeted image distribution
		invariant (or approximately invariant) and if $p$ is the distribution of a pretrained model, then we form the ensemble
		$\frac{1}{K} \sum_k p(\mathbf{T}_k \mathbf{x}) |\det \mathbf{T}_k|$. Note that this is simply a mixture
		model over images $\mathbf{x}$. We considered rotating as well as flipping images along the horizontal and
		vertical axes (yielding an ensemble over 8 transformations). While it could be argued that most
		of these transformations do not leave the distribution over natural images invariant, we
		nevertheless observed a noticeable boost in performance.

		\subsection{Natural images}
		\begin{figure}[t]
			\begin{floatrow}
				\capbtabbox{
					\small
					\setlength{\tabcolsep}{.1em}
					\begin{tabular}{lccc}
						\hline
						\textbf{Model} &
						\parbox{1.1cm}{\centering\textbf{63 dim.} \\ \textbf{[nat]}} &
						\parbox{1.1cm}{\centering\textbf{64 dim.} \\ \textbf{[bit/px]}} &
						\parbox{1.1cm}{\centering\textbf{$\mathbf{\infty}$ dim.} \\ \textbf{[bit/px]}} \\
						\hline
						RNADE \cite{Uria:2013}                            & 152.1          & \cellcolor{gray!20} 3.346 & - \\
						RNADE, 1 hl \cite{Uria:2014}                      & 143.2          & \cellcolor{gray!20} 3.146 & - \\
						RNADE, 6 hl \cite{Uria:2014}                      & 155.2          & \cellcolor{gray!20} 3.416 & - \\
						EoRNADE, 6 layers \cite{Uria:2014}                & 157.0          & \cellcolor{gray!20} 3.457 & - \\
						GMM, 200 comp. \cite{Zoran:2012,VanDenOord:2014b} & 153.7          & \cellcolor{gray!20} 3.360 & - \\
						STM, 200 comp. \cite{VanDenOord:2014a}            & 155.3          & \cellcolor{gray!20} 3.418 & - \\
						Deep GMM, 3 layers \cite{VanDenOord:2014b}        & 156.2          & \cellcolor{gray!20} 3.439 & - \\
						MCGSM, 16 comp.                                   & 155.1          & 3.413                     & \cellcolor{gray!20} 3.688 \\
						MCGSM, 32 comp.                                   & 155.8          & 3.430                     & \cellcolor{gray!20} 3.706 \\
						MCGSM, 64 comp.                                   & 156.2          & 3.439                     & \cellcolor{gray!20} 3.716 \\
						MCGSM, 128 comp.                                  & 156.4          & 3.443                     & \cellcolor{gray!20} 3.717 \\
						EoMCGSM, 128 comp.                                & \textbf{158.1} & \textbf{3.481}            & \cellcolor{gray!20} 3.748 \\
						RIDE, 1 layer                                     & 150.7          & 3.293                     & \cellcolor{gray!20} 3.802  \\
						RIDE, 2 layers                                    & 152.1          & 3.346                     & \cellcolor{gray!20} 3.869 \\
						EoRIDE, 2 layers                                  & 154.5          & 3.400                     & \cellcolor{gray!20} \textbf{3.899} \\
						\hline
					\end{tabular}
				}{
					\caption{Average log-likelihoods and log-likelihood rates for image
						patches (without/with DC comp.) and large images extracted from BSDS300
						\cite{Martin:2001}.}
					\label{tbl:bsds300}
				}
				\capbtabbox{
					\small
					\setlength{\tabcolsep}{.1em}
					\begin{tabular}{lcc}
						\hline
						\textbf{Model} &
						\parbox{1.2cm}{\centering\textbf{256 dim.} \\ \textbf{[bit/px]}} &
						\parbox{1.2cm}{\centering\textbf{$\mathbf{\infty}$ dim.} \\ \textbf{[bit/px]}} \\
						\hline
						GRBM \cite{Hinton:2006}                & \cellcolor{gray!20} 0.992 & - \\
						ICA \cite{Bell:1997,vanHateren:1998}   & \cellcolor{gray!20} 1.072 & - \\
						GSM                                    & \cellcolor{gray!20} 1.349 & - \\
						ISA \cite{Hyvarinen:2000,Gerhard:2015} & \cellcolor{gray!20} 1.441 & - \\
						MoGSM, 32 comp. \cite{Theis:2011}      & \cellcolor{gray!20} 1.526 & - \\
						MCGSM, 32 comp.                        & 1.615                     & \cellcolor{gray!20} 1.759 \\
						RIDE, 1 layer, 64 hid.                 & \textbf{1.650}            & \cellcolor{gray!20} 1.816 \\
						RIDE, 1 layer, 128 hid.                & -                         & \cellcolor{gray!20} 1.830 \\
						RIDE, 2 layers, 64 hid.                & -                         & \cellcolor{gray!20} 1.829 \\
						RIDE, 2 layers, 128 hid.               & -                         & \cellcolor{gray!20} 1.839  \\
						EoRIDE, 2 layers, 128 hid.             & -                         & \cellcolor{gray!20} \textbf{1.859}  \\
						\hline
					\end{tabular}
				}{
					\caption{Average log-likelihood rates for image patches and large
						images extracted from van Hateren's dataset \cite{vanHateren:1998}.}
					\label{tbl:vanhateren}
				}
			\end{floatrow}
		\end{figure}
		Several recent image models have been evaluated on small image patches sampled from
		the Berkeley segmentation dataset (BSDS300) \cite{Martin:2001}. Although our model's
		strength lies in its ability to scale to large images and to capture long-range correlations, we include
		results on BSDS300 to make a connection to this part of the literature. We followed the
		protocol of Uria et al. \cite{Uria:2013}. The RGB images were turned to
		grayscale, uniform noise was added to account for the integer discretization, and the resulting
		values were divided by 256. The training set of 200 images was split into 180 images for
		training and 20 images for validation, while the test set contained 100 images. We extracted 8 by 8
		image patches from each set and subtracted the average pixel intensity such
		that each patch's DC component was zero. Because the resulting image patches live on a 63
		dimensional subspace, the bottom-right pixel was discarded. We used
		$1.6 \cdot 10^6$ patches for training, $1.8 \cdot 10^5$ patches for validation, and
		$10^6$ test patches for evaluation.

		MCGSMs have not been evaluated on this dataset and so we first tested MCGSMs by training a single factorized MCGSM
		for each pixel conditioned on all previous pixels in a fixed ordering. We find that already
		an MCGSM (with 128 components and 48 quadratic features) outperforms all single models
		including a deep Gaussian mixture model \cite{VanDenOord:2014a} (Table~\ref{tbl:bsds300}).
		Our ensemble of MCGSMs\footnote{Details on how the
		ensemble of transformations can be applied despite the missing bottom-right pixel are given
		in Supplementary Section 2.1.} outperforms an ensemble of RNADEs with 6 hidden layers, which to
		our knowledge is currently the best result reported on this dataset.

		Training the recurrent image density estimator (RIDE) on the 63 dimensional dataset is more cumbersome. We tried
		padding image patches with zeros, which was necessary
		to be able to compute a hidden state at every pixel. The bottom-right pixel was ignored during training and evaluation. This simple approach led to a reduction in
		performance relative to the MCGSM (Table~\ref{tbl:bsds300}). A possible explanation is that
		the model cannot distinguish between pixel intensities which are zero and zeros in the padded region.
		Supplying the model with additional binary indicators as inputs (one for each neighborhood pixel) did
		not solve the problem.

		However, we found that RIDE outperforms the MCGSM by a large margin when images were treated
		as instances of a stochastic process (that is, using infinitely large images). MCGSMs were trained for up to 3000
		iterations of L-BFGS on $10^6$ pixels and corresponding causal neighborhoods extracted from the training images.
		Causal neighborhoods were 9 pixels wide and 5 pixels high. RIDE was trained
		for 8 epochs on image patches of increasing size ranging from 8 by 8 to 22 by 22 pixels
		(that is, gradients were approximated as in backpropagation through time \cite{Robinson:1987}). The right column in
		Table~\ref{tbl:bsds300} shows average log-likelihood rates for both models. Analogously to the entropy rate
		\cite{Cover:2006}, we have for the expected log-likelihood rate:
		\begin{align}
			\lim_{N \rightarrow \infty} \mathbb{E}\left[ \log p(\mathbf{x})/N^2 \right]
			&= \mathbb{E}[\log p(x_{ij} \mid \mathbf{x}_{<ij})],
		\end{align}
		where $\mathbf{x}$ is an $N$ by $N$ image patch. An average log-likelihood rate can be directly
		computed for the MCGSM, while for RIDE and ensembles we approximated it by splitting the test images into
		64 by 64 patches and evaluating on those.
		\begin{figure}[t]
			\vspace{-.3cm}
			\begin{floatrow}
				\capbtabbox[6.8cm]{
					\setlength{\tabcolsep}{.12em}
					\small
					\begin{tabular}{lc}
						\hline
						\textbf{Model} & \textbf{[bit/px]} \\
						\hline
						MCGSM, 12 comp. \cite{Theis:2012a} & 1.244 \\
						MCGSM, 32 comp. & 1.294 \\
						Diffusion \cite{Sohl-Dickstein:2015} & 1.489 \\
						RIDE, 64 hid., 1 layer & 1.402 \\
						RIDE, 64 hid., 1 layer, ext. & 1.416 \\
						RIDE, 64 hid., 2 layers & 1.438 \\
						RIDE, 64 hid., 3 layers & 1.454 \\
						RIDE, 128 hid., 3 layers & 1.489 \\
						EoRIDE, 128 hid., 3 layers & \textbf{1.501} \\
						\hline
					\end{tabular}
					\vspace{.2cm}
				}{
					\caption{Average log-likelihood rates on dead leaf images. A deep recurrent image model
					is on a par with a deep diffusion model \cite{Sohl-Dickstein:2015}. Using
					ensembles we are able to further improve the likelihood.}
					\label{tbl:deadleaves}
				}
				\ffigbox[6.8cm]{
					\hspace{-1cm}
					\begin{tikzpicture}
	\begin{axis}[
		scale only axis,
		width=3.5cm,
		height=3cm,
		at={(0.0cm, 0.0cm)},
		xmin=2,
		xmax=14,
		ymin=1,
		ymax=1.5,
		xlabel={Neighborhood size},
		ylabel={Log-likelihood [bit/px]},
		legend entries={MCGSM,RIDE},
		legend cell align=left,
		legend style={
			at={(0.98,0.02)},
			anchor=south east,
			font=\fontsize{8pt}{8pt}\sansmath\sffamily\selectfont,
		},
		xlabel near ticks,
		ylabel near ticks,
		grid=major,
		xtick={3,5,7,9,11,13},
		ytick={1,1.1,1.2,1.3,1.4,1.5},
		xticklabel={\pgfmathprintnumber[precision=4]{\tick}},
		yticklabel={\pgfmathprintnumber[precision=4]{\tick}},
		zticklabel={\pgfmathprintnumber[precision=4]{\tick}},
		tick label style={font=\footnotesize\sansmath\sffamily\selectfont},
		label style={font=\footnotesize\sansmath\sffamily\selectfont},
		axis x line=bottom,
		axis y line=left]
		\addplot+[
			no marks,
			densely dashed,
			line width=2pt,
			mark options={solid},
			color={rgb,1:red,0.90;green,0.00;blue,0.00}] coordinates {
			(3, 1.13458)
			(5, 1.2745)
			(7, 1.2948)
			(9, 1.29102)
			(11, 1.29408)
			(13, 1.28352)
		};
		\addplot+[
			no marks,
			solid,
			line width=2pt,
			mark options={solid},
			color={rgb,1:red,0.10;green,0.60;blue,1.00}] coordinates {
			(3, 1.388)
			(5, 1.402)
			(7, 1.38)
			(9, 1.374)
		};
	\end{axis}
\end{tikzpicture}
					\vspace{-.2cm}
				}{
					\caption{Model performance on dead leaves as a function of the causal
						neighborhood width. Simply increasing the neighborhood size of the MCGSM is not sufficient to
						improve performance.}
					\label{fig:nb_size}
				}
			\end{floatrow}
		\end{figure}

		To make the two sets of numbers more comparable, we transformed nats as commonly reported on the
		63 dimensional data, $\ell_{1:63}$, into a bit per pixel log-likelihood rate using the formula
		$(\ell_{1:63} + \ell_{DC} + \ln |\det \mathbf{A}|) / 64 / \ln(2)$. This takes into account a log-likelihood
		for the missing DC component, $\ell_{DC} = 0.5020$, and the Jacobian of the transformations applied during
		preprocessing, $\ln |\det \mathbf{A}| = -4.1589$ (see Supplementary Section
		2.2 for details). The two rates in Table~\ref{tbl:bsds300} are comparable in the sense that their differences express
		how much better one model would be at losslessly compressing BSDS300 test images than another, where patch-based models
		would compress patches of an image independently. We highlighted the best result
		achieved with each model in gray. Note that most models in this list do not scale as well to
		large images as the MCGSM or RIDE (GMMs in particular) and are therefore unlikely to benefit
		as much from increasing the patch size.

		A comparison of the log-likelihood rates reveals that an MCGSM with 16 components applied to
		large images already captures more correlations than any model applied to small image patches.
		The difference is particularly striking given that the factorized MCGSM has approximately 3,000
		parameters while a GMM with 200 components has approximately 400,000 parameters.
		Using an ensemble of RIDEs, we are able to further improve this number significantly
		(Table~\ref{tbl:bsds300}).

		Another dataset frequently used to test generative image models is the dataset published by
		van Hateren and van der Schaaf \cite{vanHateren:1998}. Details of the preprocessing used in
		this paper are given in Supplementary Section~3. We reevaluated several models for which the
		likelihood has been reported on this dataset \cite{Gerhard:2015,Theis:2011,Theis:2012a,Theis:2012b}.
		Likelihood rates as well as results on 16 by 16 patches are given in Table~\ref{tbl:vanhateren}.
		Because of the larger patch size, RIDE here already outperforms the MCGSM on patches.

		\subsection{Dead leaves}
		Dead leaf images are generated by superimposing disks of random intensity and
		size on top of each other \cite{Matheron:1968,Lee:2001}. This simple procedure leads to images which already share many
		of the statistical properties and challenges of natural images, such as occlusions and
		long-range correlations, while leaving out others such as non-stationary statistics. They
		therefore provide an interesting test case for natural image models.

		We used a set of 1,000 images, where each image is 256 by 256 pixels in size.
		We compare the performance of RIDE to the MCGSM and a very recently introduced deep
		multiscale model based on a diffusion process \cite{Sohl-Dickstein:2015}.
		The same 100 images as in previous literature \cite{Theis:2012a,Sohl-Dickstein:2015} were
		used for evaluation and we used the remaining images for training. We find that the
		introduction of an SLSTM with 64 hidden units greatly improves the performance of the MCGSM. We also tried an extended version of
		the SLSTM which included memory units as additional inputs (right-hand side of Equation~\ref{eq:slstm}).
		This yielded a small improvement in performance (5th row in Table~\ref{tbl:deadleaves}) while adding layers or using more
		hidden units led to more drastic improvements. Using 3 layers with 128 hidden units in each
		layer, we find that our recurrent image model is on a par with the deep diffusion model.
		By using ensembles, we are able to beat all previously published results for this dataset
		(Table~\ref{tbl:deadleaves}).

		Figure~\ref{fig:nb_size} shows that the improved performance of RIDE is not simply due to an
		effectively larger causal neighborhood but that the nonlinear transformations performed by
		the SLSTM units matter. Simply increasing the neighborhood size of an MCGSM does not yield
		the same improvement. Instead, the performance quickly saturates. We also find that the performance
		of RIDE slightly deteriorates with larger neighborhoods, which is likely caused by optimization
		difficulties.
		\begin{figure}[t]
			\vspace{-.3cm}
			\centering
			\input{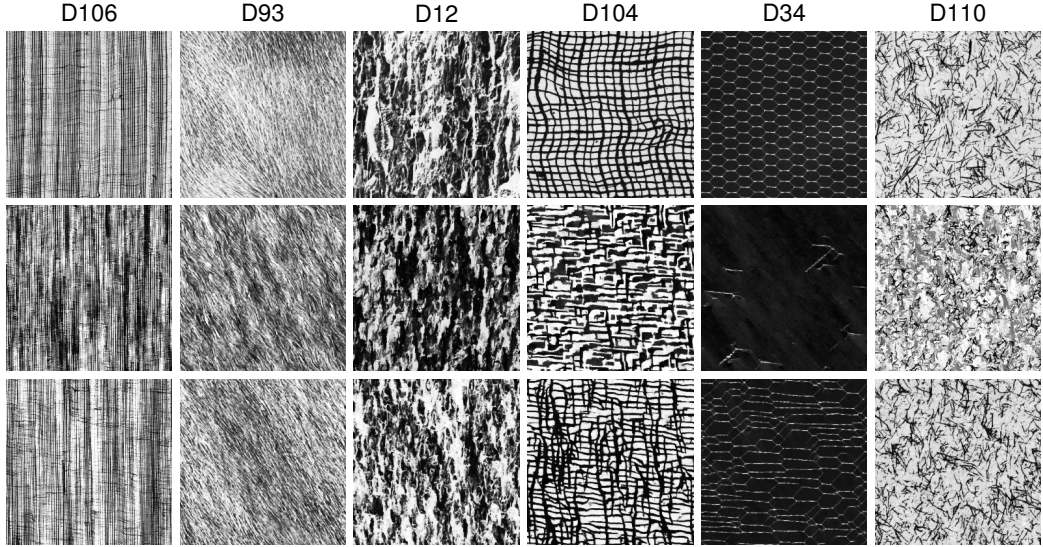}
			\vspace{-.25cm}
			\caption{From top to bottom: A 256 by 256 pixel crop of
				a texture \cite{Brodatz:1966}, a sample generated by an MCGSM trained on the
				full texture \cite{Gerhard:2015}, and a sample generated by RIDE. This
				illustrates that our model can capture a variety of different statistical
				patterns. The addition of the recurrent neural network seems particularly
			helpful where there are strong long-range correlations (D104, D34).}
			\label{fig:textures}
		\end{figure}

		\subsection{Texture synthesis and inpainting}
		To get an intuition for the kinds of correlations which RIDE
		can capture or fails to capture, we tried to use it to synthesize textures.
		We used several 640 by 640 pixel textures published by Brodatz
		\cite{Brodatz:1966}. The textures were split into sixteen 160 by 160 pixel regions
		of which 15 were used for training and one randomly selected region was kept for testing
		purposes. RIDE was trained for up to 6 epochs on patches of increasing size ranging from
		20 by 20 to 40 by 40 pixels.

		Samples generated by an MCGSM and RIDE are shown in Figure~\ref{fig:textures}.
		Both models are able to capture a wide range of correlation structures.
		However, the MCGSM seems to struggle with textures having bimodal marginal distributions
		and periodic patterns (D104, D34, and D110). RIDE clearly improves on these
		textures, although it also struggles to faithfully reproduce periodic structure. Possible
		explanations include that LSTMs are not well suited to capture periodicities, or that
		these failures are not penalized strong enough by the likelihood.
		For some textures, RIDE produces samples which are nearly indistinguishable from the real
		textures (D106 and D110).

\begin{figure}
			\vspace{-.2cm}
			\centering
			\input{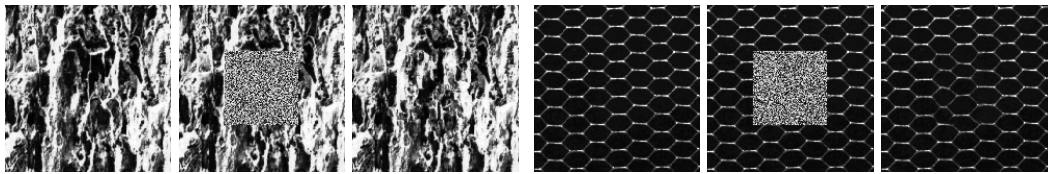}
			\vspace{-.6cm}
			\caption{The center portion of a texture (left and center) was reconstructed by sampling from the
			posterior distribution of RIDE (right).}
			\label{fig:inpainting}
		\end{figure}

		One application of generative image models is inpainting \cite[e.g.,][]{Roth:2009,Heess:2009,Sohl-Dickstein:2015}.
		As a proof of concept, we used our model to inpaint a large (here, 71 by 71 pixels) region
		in textures (Figure~\ref{fig:inpainting}). Missing pixels were replaced by sampling from the posterior of RIDE.
		Unlike the joint distribution, the posterior distribution cannot be sampled directly and we had to resort to Markov
		chain Monte Carlo methods. We found the following \textit{Metropolis within
		Gibbs} \cite{Tierney:1994} procedure to be efficient enough. The missing pixels were initialized via ancestral sampling. Since
		ancestral sampling is cheap, we generated 5 candidates and used the one with
		the largest posterior density. Following initialization, we sequentially updated
		overlapping 5 by 5 pixel regions via Metropolis sampling. Proposals were generated
		via ancestral sampling and accepted using the acceptance probability
		\begin{align}
			\textstyle\alpha = \min\left\{ 1, \frac{p(\mathbf{x}')}{p(\mathbf{x})} \frac{p(\mathbf{x}_{ij} \mid \mathbf{x}_{<ij})}{p(\mathbf{x}_{ij}' \mid \mathbf{x}_{<ij})} \right\},
		\end{align}
		where here $\mathbf{x}_{ij}$ represents a 5 by 5 pixel patch and $\mathbf{x}_{ij}'$ its
		proposed replacement. Since evaluating the joint and conditional densities on the entire image is
		costly, we approximated $p$ using RIDE applied to a 19 by 19 pixel patch surrounding $ij$.
		Randomly flipping images vertically or horizontally in between the sampling further helped.
		Figure~\ref{fig:inpainting} shows results after 100 Gibbs sampling sweeps.
		\vspace{-.15cm}

	\section{Conclusion}
		We have introduced RIDE, a deep but tractable recurrent image model based on spatial LSTMs.
		The model exemplifies how recent insights in deep learning can be exploited for generative
		image modeling and shows superior performance in quantitative comparisons. RIDE is able to capture many
		different statistical patterns, as demonstrated through its application to textures. This is an
		important property considering that on an intermediate level of abstraction natural images
		can be viewed as collections of textures.

		We have furthermore introduced a factorized version of
		the MCGSM which allowed us to use more experts and larger causal neighborhoods. This
		model has few parameters, is easy to train and already on its own performs very
		well as an image model. It is therefore an ideal building block and may be used to extend other models such as DRAW
		\cite{Gregor:2015} or video models \cite{Ranzato:2014,Srivastava:2015}.

		Deep generative image models have come a long way since deep belief
		networks have first been applied to natural images \cite{Osindero:2008}. Unlike convolutional neural networks in object recognition,
		however, no approach has as of yet proven to be a likely solution to the problem of
		generative image modeling. Further conceptual work will be necessary to come up with a model
		which can handle both the more abstract high-level as well as the low-level statistics of natural images.

	\subsubsection*{Acknowledgments}
		The authors would like to thank Aäron van den Oord for insightful discussions and Wieland Brendel,
		Christian Behrens, and Matthias Kümmerer for helpful input on this paper. This study was
		financially supported by the German Research Foundation (DFG; priority program 1527, BE 3848/2-1).

	\bibliographystyle{plainnat}
	\bibliography{references}

\begin{thebibliography}{50}
\providecommand{\natexlab}[1]{#1}
\providecommand{\url}[1]{\texttt{#1}}
\expandafter\ifx\csname urlstyle\endcsname\relax
  \providecommand{\doi}[1]{doi: #1}\else
  \providecommand{\doi}{doi: \begingroup \urlstyle{rm}\Url}\fi

\bibitem[Bell and Sejnowski(1997)]{Bell:1997}
A.~J. Bell and T.~J. Sejnowski.
\newblock The ``independent components'' of natural scenes are edge filters.
\newblock \emph{Vision Research}, 37\penalty0 (23):\penalty0 3327--3338, 1997.

\bibitem[Brodatz(1966)]{Brodatz:1966}
P.~Brodatz.
\newblock \emph{Textures: A Photographic Album for Artists and Designers}.
\newblock Dover, New York, 1966.
\newblock URL \url{http://www.ux.uis.no/~tranden/brodatz.html}.

\bibitem[Cover and Thomas(2006)]{Cover:2006}
T.~Cover and J.~Thomas.
\newblock \emph{Elements of Information Theory}.
\newblock Wiley, 2nd edition, 2006.

\bibitem[Denton et~al.(2015)Denton, Chintala, Szlam, and Fergus]{Denton:2015}
E.~Denton, S.~Chintala, A.~Szlam, and R.~Fergus.
\newblock {Deep Generative Image Models using a {Laplacian} Pyramid of
  Adversarial Networks}.
\newblock In \emph{Advances in Neural Information Processing Systems 28}, 2015.

\bibitem[Domke et~al.(2008)Domke, Karapurkar, and Aloimonos]{Domke:2008}
J.~Domke, A.~Karapurkar, and Y.~Aloimonos.
\newblock Who killed the directed model?
\newblock In \emph{CVPR}, 2008.

\bibitem[Donahue et~al.(2014)Donahue, Jia, Vinyals, Hoffman, Zhang, Tzeng, and
  Darrell]{Donahue:2014}
J.~Donahue, Y.~Jia, O.~Vinyals, J.~Hoffman, N.~Zhang, E.~Tzeng, and T.~Darrell.
\newblock {DeCAF}: A deep convolutional activation feature for generic visual
  recognition.
\newblock In \emph{ICML 31}, 2014.

\bibitem[Gerhard et~al.(2015)Gerhard, Theis, and Bethge]{Gerhard:2015}
H.~E. Gerhard, L.~Theis, and M.~Bethge.
\newblock Modeling natural image statistics.
\newblock In \emph{Biologically-inspired Computer Vision—Fundamentals and
  Applications}. Wiley VCH, 2015.

\bibitem[Goodfellow et~al.(2014)Goodfellow, Pouget-Abadie, Mirza, Xu,
  Warde-Farley, Ozair, Courville, and Bengio]{Goodfellow:2014}
I.~Goodfellow, J.~Pouget-Abadie, M.~Mirza, B.~Xu, D.~Warde-Farley, S.~Ozair,
  A.~Courville, and Y.~Bengio.
\newblock Generative adversarial nets.
\newblock In \emph{Advances in Neural Information Processing Systems 27}, 2014.

\bibitem[Graves and Schmidhuber(2009)]{Graves:2009}
A.~Graves and J.~Schmidhuber.
\newblock Offline handwriting recognition with multidimensional recurrent
  neural networks.
\newblock In \emph{Advances in Neural Information Processing Systems 22}, 2009.

\bibitem[Gregor et~al.(2014)Gregor, Danihelka, Mnih, Blundell, and
  Wierstra]{Gregor:2014}
K.~Gregor, I.~Danihelka, A.~Mnih, C.~Blundell, and D.~Wierstra.
\newblock {Deep AutoRegressive Networks}.
\newblock In \emph{Proceedings of the 31st International Conference on Machine
  Learning}, 2014.

\bibitem[Gregor et~al.(2015)Gregor, Danihelka, Graves, and
  Wierstra]{Gregor:2015}
K.~Gregor, I.~Danihelka, A.~Graves, and D.~Wierstra.
\newblock {DRAW}: A recurrent neural network for image generation.
\newblock In \emph{Proceedings of the 32nd International Conference on Machine
  Learning}, 2015.

\bibitem[Heess et~al.(2009)Heess, Williams, and Hinton]{Heess:2009}
N.~Heess, C.~Williams, and G.~E. Hinton.
\newblock Learning generative texture models with extended fields-of-experts.
\newblock In \emph{BMCV}, 2009.

\bibitem[Hinton et~al.(2006)Hinton, Osindero, and Teh]{Hinton:2006}
G.~Hinton, S.~Osindero, and Y.~Teh.
\newblock A fast learning algorithm for deep belief nets.
\newblock \emph{Neural Comp.}, 2006.

\bibitem[Hochreiter and Schmidhuber(1997)]{Hochreiter:1997}
S.~Hochreiter and J.~Schmidhuber.
\newblock Long short-term memory.
\newblock \emph{Neural Computation}, 9\penalty0 (8), 1997.

\bibitem[Hosseini et~al.(2010)Hosseini, Sinz, and Bethge]{Hosseini:2010}
R.~Hosseini, F.~Sinz, and M.~Bethge.
\newblock Lower bounds on the redundancy of natural images.
\newblock \emph{Vis. Res.}, 2010.

\bibitem[Hyvärinen and Hoyer(2000)]{Hyvarinen:2000}
A.~Hyvärinen and P.~O. Hoyer.
\newblock Emergence of phase and shift invariant features by decomposition of
  natural images into independent feature subspaces.
\newblock \emph{Neural Computation}, 12\penalty0 (7):\penalty0 1705–--1720,
  2000.

\bibitem[Jia et~al.(2014)Jia, Shelhamer, Donahue, Karayev, Long, Girshick,
  Guadarrama, and Darrell]{jia:2014}
Y.~Jia, E.~Shelhamer, J.~Donahue, S.~Karayev, J.~Long, R.~Girshick,
  S.~Guadarrama, and T.~Darrell.
\newblock Caffe: Convolutional architecture for fast feature embedding, 2014.
\newblock arXiv:1408.5093.

\bibitem[Kingma and Welling(2014)]{Kingma:2014a}
D.~P. Kingma and M.~Welling.
\newblock Auto-encoding variational {Bayes}.
\newblock In \emph{ICLR}, 2014.

\bibitem[Kingma et~al.(2014)Kingma, Rezende, Mohamed, and
  Welling]{Kingma:2014b}
D.~P. Kingma, D.~J. Rezende, S.~Mohamed, and M.~Welling.
\newblock Semi-supervised learning with deep generative models.
\newblock In \emph{Advances in Neural Information Processing Systems 27}, 2014.

\bibitem[Krizhevsky et~al.(2012)Krizhevsky, Sutskever, and
  Hinton]{Krizhevsky:2012}
A.~Krizhevsky, I.~Sutskever, and G.~E. Hinton.
\newblock {ImageNet} classification with deep convolutional neural networks.
\newblock In \emph{Advances in Neural Information Processing Systems 25}, 2012.

\bibitem[Larochelle and Murray(2011)]{Larochelle:2011}
H.~Larochelle and I.~Murray.
\newblock The neural autoregressive distribution estimator.
\newblock In \emph{Proceedings of the 14th International Conference on
  Artificial Intelligence and Statistics}, 2011.

\bibitem[Lee et~al.(2001)Lee, Mumford, and Huang]{Lee:2001}
A.~B. Lee, D.~Mumford, and J.~Huang.
\newblock Occlusion models for natural images: A statistical study of a
  scale-invariant dead leaves model.
\newblock \emph{International Journal of Computer Vision}, 2001.

\bibitem[Lee et~al.(2009)Lee, Grosse, Ranganath, and Ng]{Lee:2009}
H.~Lee, R.~Grosse, R.~Ranganath, and A.~Y. Ng.
\newblock Convolutional deep belief networks for scalable unsupervised learning
  of hierarchical representations.
\newblock In \emph{ICML 26}, 2009.

\bibitem[Li et~al.(2015)Li, Swersky, and Zemel]{Li:2015}
Y.~Li, K.~Swersky, and R.~Zemel.
\newblock Generative moment matching networks.
\newblock In \emph{ICML 32}, 2015.

\bibitem[Martin et~al.(2001)Martin, Fowlkes, Tal, and Malik]{Martin:2001}
D.~Martin, C.~Fowlkes, D.~Tal, and J.~Malik.
\newblock A database of human segmented natural images and its application to
  evaluating segmentation algorithms and measuring ecological statistics.
\newblock In \emph{ICCV}, 2001.

\bibitem[Matheron(1968)]{Matheron:1968}
G.~Matheron.
\newblock Modele s\'{e}quential de partition al\'{e}atoire.
\newblock Technical report, CMM, 1968.

\bibitem[Neal(1992)]{Neal:1992}
R.~M. Neal.
\newblock Connectionist learning of belief networks.
\newblock \emph{Artificial Intelligence}, 56:\penalty0 71--113, 1992.

\bibitem[Ngiam et~al.(2011)Ngiam, Chen, Koh, and Ng]{Ngiam:2011}
J.~Ngiam, Z.~Chen, P.~W. Koh, and A.~Y. Ng.
\newblock Learning deep energy models.
\newblock In \emph{ICML 28}, 2011.

\bibitem[Osindero and Hinton(2008)]{Osindero:2008}
S.~Osindero and G.~E. Hinton.
\newblock Modelling image patches with a directed hierarchy of markov random
  fields.
\newblock In \emph{Advances In Neural Information Processing Systems 20}, 2008.

\bibitem[Ranzato et~al.(2011)Ranzato, Susskind, Mnih, and Hinton]{Ranzato:2011}
M.~A. Ranzato, J.~Susskind, V.~Mnih, and G.~E. Hinton.
\newblock On deep generative models with applications to recognition.
\newblock In \emph{IEEE Conference on Computer Vision and Pattern Recognition},
  2011.

\bibitem[Ranzato et~al.(2015)Ranzato, Szlam, Bruna, Mathieu, Collobert, and
  Chopra]{Ranzato:2014}
M.~A. Ranzato, A.~Szlam, J.~Bruna, M.~Mathieu, R.~Collobert, and S.~Chopra.
\newblock Video (language) modeling: a baseline for generative models of
  natural videos, 2015.
\newblock arXiv:1412.6604v2.

\bibitem[Robinson and Fallside(1987)]{Robinson:1987}
A.~J. Robinson and F.~Fallside.
\newblock The utility driven dynamic error propagation network.
\newblock Technical report, Cambridge University, 1987.

\bibitem[Roth and Black(2009)]{Roth:2009}
S.~Roth and M.~J. Black.
\newblock Fields of experts.
\newblock \emph{International Journal of Computer Vision}, 82\penalty0 (2),
  2009.

\bibitem[Simonyan and Zisserman(2015)]{Simonyan:2015}
K.~Simonyan and A.~Zisserman.
\newblock Very deep convolutional networks for large-scale image recognition.
\newblock In \emph{International Conference on Learning Represenations}, 2015.

\bibitem[Sohl-Dickstein et~al.(2015)Sohl-Dickstein, Weiss, Maheswaranathan, and
  Ganguli]{Sohl-Dickstein:2015}
J.~Sohl-Dickstein, E.~A. Weiss, N.~Maheswaranathan, and S.~Ganguli.
\newblock Deep unsupervised learning using nonequilibrium thermodynamics.
\newblock In \emph{ICML 32}, 2015.

\bibitem[Srivastava and Salakhutdinov(2014)]{Srivastava:2014}
N.~Srivastava and R.~Salakhutdinov.
\newblock Multimodal learning with deep {Boltzmann} machines.
\newblock \emph{JMLR}, 2014.

\bibitem[Srivastava et~al.(2015)Srivastava, Mansimov, and
  Salakhutdinov]{Srivastava:2015}
N.~Srivastava, E.~Mansimov, and R.~Salakhutdinov.
\newblock Unsupervised learning of video representations using {LSTMs}.
\newblock In \emph{Proceedings of the 32nd International Conference on Machine
  Learning}, 2015.

\bibitem[Sundermeyer et~al.(2010)Sundermeyer, Schluter, and
  Ney]{Sundermeyer:2010}
M.~Sundermeyer, R.~Schluter, and H.~Ney.
\newblock {LSTM} neural networks for language modeling.
\newblock In \emph{INTERSPEECH}, 2010.

\bibitem[Sutskever et~al.(2014)Sutskever, Vinyals, and Le]{Sutskever:2014}
I.~Sutskever, O.~Vinyals, and Q.~V. Le.
\newblock Sequence to sequence learning with neural networks.
\newblock In \emph{Advances in Neural Information Processing Systems 27}, 2014.

\bibitem[Theis et~al.(2011)Theis, Gerwinn, Sinz, and Bethge]{Theis:2011}
L.~Theis, S.~Gerwinn, F.~Sinz, and M.~Bethge.
\newblock In all likelihood, deep belief is not enough.
\newblock \emph{JMLR}, 2011.

\bibitem[Theis et~al.(2012{\natexlab{a}})Theis, Hosseini, and
  Bethge]{Theis:2012a}
L.~Theis, R.~Hosseini, and M.~Bethge.
\newblock Mixtures of conditional {Gaussian} scale mixtures applied to
  multiscale image representations.
\newblock \emph{PLoS ONE}, 7\penalty0 (7), 2012{\natexlab{a}}.

\bibitem[Theis et~al.(2012{\natexlab{b}})Theis, Sohl-Dickstein, and
  Bethge]{Theis:2012b}
L.~Theis, J.~Sohl-Dickstein, and M.~Bethge.
\newblock Training sparse natural image models with a fast {Gibbs} sampler of
  an extended state space.
\newblock In \emph{Advances in Neural Information Processing Systems 25},
  2012{\natexlab{b}}.

\bibitem[Tierney(1994)]{Tierney:1994}
L.~Tierney.
\newblock Markov chains for exploring posterior distributions.
\newblock \emph{The Annals of Statistics}, 1994.

\bibitem[Uria et~al.(2013)Uria, Murray, and Larochelle]{Uria:2013}
B.~Uria, I.~Murray, and H.~Larochelle.
\newblock {RNADE:} the real-valued neural autoregressive density-estimator.
\newblock In \emph{Advances in Neural Information Processing Systems 26}, 2013.

\bibitem[Uria et~al.(2014)Uria, Murray, and Larochelle]{Uria:2014}
B.~Uria, I.~Murray, and H.~Larochelle.
\newblock A deep and tractable density estimator.
\newblock In \emph{ICML 31}, 2014.

\bibitem[van~den Oord and Schrauwen(2014{\natexlab{a}})]{VanDenOord:2014a}
A.~van~den Oord and B.~Schrauwen.
\newblock {The student-t mixture as a natural image patch prior with
  application to image compression.}
\newblock \emph{Journal of Machine Learning Research}, 15\penalty0
  (1):\penalty0 2061--2086, 2014{\natexlab{a}}.

\bibitem[van~den Oord and Schrauwen(2014{\natexlab{b}})]{VanDenOord:2014b}
A.~van~den Oord and B.~Schrauwen.
\newblock Factoring variations in natural images with deep {Gaussian} mixture
  models.
\newblock In \emph{Advances in Neural Information Processing Systems 27},
  2014{\natexlab{b}}.

\bibitem[van Hateren and van~der Schaaf(1998)]{vanHateren:1998}
J.~H. van Hateren and A.~van~der Schaaf.
\newblock Independent component filters of natural images compared with simple
  cells in primary visual cortex.
\newblock \emph{Proc. of the Royal Society B: Biological Sciences},
  265\penalty0 (1394), 1998.

\bibitem[Zoran and Weiss(2011)]{Zoran:2011}
D.~Zoran and Y.~Weiss.
\newblock From learning models of natural image patches to whole image
  restoration.
\newblock In \emph{IEEE International Conference on Computer Vision}, 2011.

\bibitem[Zoran and Weiss(2012)]{Zoran:2012}
D.~Zoran and Y.~Weiss.
\newblock Natural images, {Gaussian} mixtures and dead leaves.
\newblock In \emph{NIPS 25}, 2012.

\end{thebibliography}
\end{document}